\pdfoutput=1

\documentclass[11pt]{article}

\usepackage{listings}
\usepackage{xcolor}
\usepackage{booktabs}
\usepackage{multirow}
\usepackage{microtype}
\usepackage{amsmath}
\usepackage{amssymb}
\usepackage{graphicx}
\usepackage[normalem]{ulem}
\useunder{\uline}{\ul}{}
\usepackage{xcolor}

\usepackage{pifont}
\definecolor{codegreen}{rgb}{0,0.6,0}
\definecolor{codegray}{rgb}{0.5,0.5,0.5}
\definecolor{codepurple}{rgb}{0.58,0,0.82}
\definecolor{backcolour}{rgb}{0.95,0.95,0.92}

\usepackage{color, colortbl}
\usepackage[first=0,last=9]{lcg}

\definecolor{LightYellow}{rgb}{0.99, 0.99, 0.59}
\definecolor{LightRed}{rgb}{0.97, 0.51, 0.47}
\definecolor{LightBlue}{rgb}{0.99, 0.59, 0.99}
\definecolor{LightGreen}{rgb}{0.59, 0.99, 0.99}

\definecolor{codegreen}{rgb}{0,0.6,0}
\definecolor{codegray}{rgb}{0.5,0.5,0.5}

\definecolor{backcolour}{RGB}{245,248,250}
\definecolor{emph}{RGB}{166,88,53}
\definecolor{nightblue}{RGB}{9,49,105}
\definecolor{keywords}{RGB}{207,33,46}
\definecolor{lightpurple}{RGB}{130,81,223}

\lstdefinestyle{mystyle}{
    backgroundcolor=\color{backcolour},   
    commentstyle=\color{codegreen},
    keywordstyle=\color{keywords},
    stringstyle=\color{nightblue},
    basicstyle=\ttfamily\footnotesize,
    breakatwhitespace=false,         
    breaklines=true,                 
    captionpos=b,                    
    keepspaces=true,                 
    showspaces=false,                
    showstringspaces=false,
    showtabs=false,                  
    tabsize=2,
    frame=shadowbox,
    emph={AutoTokenizer,AutoModelForSequenceClassification,Explainer},
    emphstyle={\color{emph}},
    emph={[2]from_pretrained,compute_table},
    emphstyle={[2]\color{lightpurple}}
}

\usepackage[preprint]{acl}

\usepackage{times}
\usepackage{latexsym}

\usepackage[T1]{fontenc}

\usepackage[utf8]{inputenc}

\usepackage{microtype}

\usepackage{inconsolata}

\title{JORA: JAX Tensor-Parallel LoRA Library for Retrieval Augmented Fine-Tuning}

\author{Anique Tahir \\
  Arizona State University \\
  699 S. Mill Avenue \\
  Tempe, AZ \\
  \texttt{research@anique.org} \\\And
  Lu Cheng \\
  University of Illinois Chicago \\
  851 S. Morgan St. \\
  Chicago, IL \\
  \texttt{lucheng@uic.edu} \\\And
  Huan Liu \\
  Arizona State University \\
  699 S. Mill Avenue \\
  Tempe, AZ \\
  \texttt{huanliu@asu.edu} \\
}

\begin{document}
\maketitle
\begin{center}
\textbf{Open-source repository:}\\ \href{https://github.com/aniquetahir/JORA}{https://github.com/aniquetahir/JORA} \\
\textbf{Supplementary video:}\\ \href{https://youtu.be/-auF_9wF2S0?si=o50HBySZjTpjtWR_}{https://youtu.be/-auF\_9wF2S0?si=o50HBySZjTpjtWR\_}
\end{center}
\begin{abstract}
The scaling of Large Language Models (LLMs) for retrieval-based tasks, particularly in Retrieval Augmented Generation (RAG), faces significant memory constraints, especially when fine-tuning extensive prompt sequences. Current open-source libraries support full-model inference and fine-tuning across multiple GPUs but fall short of accommodating the efficient parameter distribution required for retrieved context. Addressing this gap, we introduce a novel framework for PEFT-compatible fine-tuning of Llama-2 models, leveraging distributed training. Our framework uniquely utilizes JAX's just-in-time (JIT) compilation and tensor-sharding for efficient resource management, thereby enabling accelerated fine-tuning with reduced memory requirements. This advancement significantly improves the scalability and feasibility of fine-tuning LLMs for complex RAG applications, even on systems with limited GPU resources. Our experiments show more than 12x improvement in runtime compared to Hugging Face/DeepSpeed implementation with four GPUs while consuming less than half the VRAM per GPU.
\end{abstract}

\section{Introduction}
Large Language Models (LLMs) like ChatGPT~\cite{achiam2023gpt} have revolutionized the field of natural language processing, paving the way for open-source alternatives that offer more flexibility in fine-tuning. Llama-2~\cite{touvron2023llama}, a prominent LLM, exemplifies this trend, offering extensive customization at the architecture level. Alongside, Parameter Efficient Fine-Tuning (PEFT)~\cite{fu2023effectiveness} techniques like Low-Rank Adaptation have emerged, optimizing resource utilization in training these models.
Retrieval Augmented Generation (RAG)~\cite{lewis2020retrieval} is a paradigm that leverages a corpus to enrich LLM prompts with relevant context. However, when fine-tuning on retrieval-based context, the quadratic memory scaling of transformer models with prompt length poses significant challenges, especially when integrating large context sizes. The training process, which employs teacher-forcing at each step of the sequence, exacerbates memory demands, creating a bottleneck for effective LLM utilization in RAG.

Current machine learning frameworks facilitate LLM fine-tuning on distributed systems, employing model and pipeline parallelism strategies. However, these frameworks lack support for PEFT, specifically in the context of parallel training. While libraries such as DeepSpeed~\cite{rasley2020deepspeed} and Accelerate~\cite{accelerate} offer data parallelism for fine-tuning the entire model, these libraries lack support for tensor-parallel training in the PEFT setting. In addition, combining multiple libraries adds unnecessary boilerplate code to glue together dependencies required for parameter-efficient and distributed training. These libraries also require boilerplate code for configuration since they target multiple models.

To bridge this gap, we introduce JORA (JAX-based LORA), a library tailored for Llama-2 models, designed to enhance the fine-tuning process for RAG applications. Utilizing JAX's just-in-time (JIT) compilation and innovative tensor-sharding techniques, JORA not only accelerates the fine-tuning process but also significantly optimizes memory usage~\cite{jax2018github}. Our evaluations across standard training GPUs demonstrate substantial improvements in training time and memory efficiency, addressing the critical challenges of PEFT in retrieval-based training. Our library also provides valuable helpers for using instruct format datasets, merging LORA parameters, and converting fine-tuned models to Hugging Face compatible formats.
Our work makes PEFT more accessible and efficient for LLMs, particularly in resource-constrained environments. By enhancing the scalability and efficiency of LLMs in retrieval augmented fine-tuning (RAFT), JORA opens new avenues for advanced natural language processing applications.

\section{Background}
JORA introduces the concept of RAFT. This workflow employs retrieved knowledge and outcomes to create context and expected outputs. The fine-tuning process encourages the model to learn a rationale to derive the output from the knowledge. Prior related work focuses on RAG, the inference counterpart of RAFT, whose bottleneck is the sequence length used for context in the prompt. Since RAFT shares the same bottleneck, our framework focuses on adding efficiency by providing a memory-efficient and distributed backend while exposing an intuitive API. We highlight the importance of RAG and the capabilities of other libraries which aim to solve related problems. We highlight how our library fills the gap.

\subsection{Retrieval Augmented Generation}
RAG has gained significant attention in recent years, with various approaches exploring it to enhance LLM generation. The integration of dense and sparse retrievers with LLMs, as discussed in \cite{robertson2009,luan2021}, highlights the diversity in retrieval techniques used for augmenting LMs. 
\citet{chen2017}, \citet{clark2017}, and others have contributed to conditioning LMs on retrieved documents, demonstrating significant improvements in knowledge-intensive tasks \cite{lee2019, guu2020, khandelwal2020, lewis2020, izacard2020, borgeaud2022, izacard2022}. The concept of chain-of-thought prompting in combination with retrieval mechanisms, as proposed by \citet{he2022}, marks a novel approach in this domain.
The evolution of LMs into agent-like models, capable of generating queries and performing actions based on prompts, is evident in the works of \citet{thoppilan2022}, who introduced models like LaMDA. \citet{yao2022b}, \citet{komeili2021}, and \citet{shuster2022a} further explored the generation of internet search queries by LMs.

\subsection{Parallel Training Libraries}
Several open-source libraries expose an interface for multi-GPU training for LLMs. Hugging Face implementation of Transformer models allows multi-GPU inference. The Transformers library also includes a trainer. Hugging Face's Accelerate~\cite{accelerate} library is a tool designed to simplify the process of running PyTorch training scripts on different devices, including CPU, single GPU, multiple GPUs, and TPUs while supporting mixed precision and distributed settings. It offers an easy-to-use API that allows users to run their PyTorch code across any distributed configuration with minimal changes, making training and inference at scale more straightforward. DeepSpeed~\cite{rasley2020deepspeed} is an open-source optimization library for PyTorch developed by Microsoft. It is designed to accelerate the training and inference of deep learning models, mainly focusing on large-scale models. The library addresses challenges such as memory constraints and slow training times, aiming to enhance deep learning workflows' performance and efficiency. Accelerate utilizes DeepSpeed for distributed training.

JORA solves several issues with prior libraries: i) we target Llama-2 models to reduce the boilerplate required for the training process, ii) we utilize JAX's jit optimizations for training to improve training performance compared to PyTorch. iii) we provide a tensor-parallel, multi-GPU implementation of training, and iv) we provide utility functions to simplify the data loading experience, fine-tuning the model, and compatibility with the Hugging Face ecosystem.

\section{JORA Framework}
\begin{figure*}
    \centering
    \includegraphics[width=.9\textwidth]{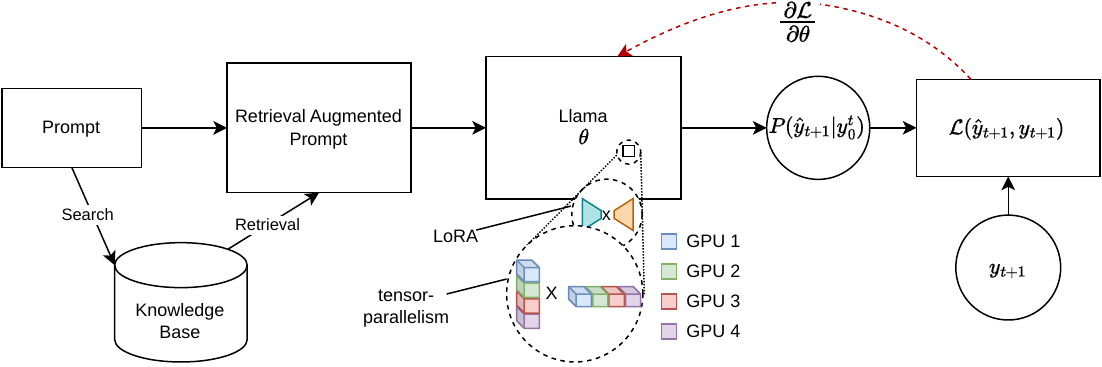}
    \caption{JORA is a library that aids in Retrieval Augmented Fine-Tuning by eliminating unnecessary boilerplate and introducing memory efficient training through tensor-parallelism and LoRA.}
    \label{fig:jora-architecture}
    \vspace{-0.5cm}
\end{figure*}
JORA is a library for RAFT. Its purpose is to make fine-tuning based on retrieved context more user-friendly. In addition, it is designed to make RAFT faster and more resource-efficient. Figure~\ref{fig:jora-architecture} gives a high-level overview of JORA. 

\subsubsection{JAX}
One of the highlights of our library is that it allows LORA training of Llama-2 models using the JAX framework. JAX provides composable transformations of numerical functions e.g. automatic differentiation~(grad), vectorization~(vmap), parallelization~(pmap), and just-in-time compilation~(jit)~\cite{jax2018github}. A function must be pure and statically composed to benefit from these transformations. Functions compiled by JAX use the Accelerated Linear Algebra (XLA) library. Jit compilation allows program optimizations to the XLA to improve execution speed which is ideal for compute-heavy architectures such as transformers. 

\subsubsection{Dataset Loading and Training}
\begin{lstlisting}[label=lst:alpaca, caption=An example of Alpaca format data.]
[
  {
    "instruction": "Calculate the area of the following shape in square centimeters.",
    "input": "rectangle of size 4 cm x 5 cm",
    "output": "20cm^2"
  },
  ...
]
\end{lstlisting}
\vspace{-0.2cm}

Even though JORA is compatible with general-purpose fine-tuning pipelines, we provide helper functions for loading training data in alpaca format~\cite{alpaca}. The Alpaca dataset format is ideal for RAFT since it follows the instruction-tuning format. Each sample in this format may contain an instruction, input~(optional), and output. Listing~\ref{lst:alpaca} shows an example of this data format. Retrieved knowledge can be used as the input and separated from the instruction and output. The output represents the sequence that the model generates. 

\begin{lstlisting}[language=python, label=lst:alpacadataset, caption=Function signature for the constructor for AlpacaDataset.]
class AlpacaDataset(Dataset):
    def __init__(self, *, path: str, split=Union[Literal['train'], Literal['test']],
                 split_percentage=0.8, tokenizer=None, max_len=512, alpaca_mix=0.3) -> None:
\end{lstlisting}
\vspace{-0.2cm}

We provide the class `AlpacaDataset' for user-friendly data loading, which inherits from PyTorch's `Dataset' class. Listing~\ref{lst:alpacadataset} shows the signature for the constructor for this class. In addition to loading the dataset, the $alpaca\_mix$ parameter allows merging a percentage of the original alpaca dataset to prevent overfitting on the fine-tuned data. The class also provides the ability to create training and testing splits based on the provided split percentage.

\subsubsection{Training API}
How fine-tuning proceeds depends on a variety of parameters. Since this library aims to simplify the training process, JORA provides common defaults for starters. In addition, it allows customization of the training process for more advanced usage. Listing~\ref{lst:config} shows the configuration class. $JAX\_PARAMS\_PATH$ specifies the location of the model parameters. $LLAMA2\_HF\_PATH$ specifies the location of Meta's model Hugging Face format. Our library uses the Hugging Face model implementation since it uses the same tokenizer. 

\begin{lstlisting}[language=python, label=lst:config, caption=JORA allows the common defaults for the configuration with room for specificity.]
class ParallamaConfig(NamedTuple):
    JAX_PARAMS_PATH: str
    LLAMA2_HF_PATH: str
    LORA_R: int = 16
    LORA_ALPHA: int = 16
    LORA_DROPOUT: float = 0.05
\end{lstlisting}
\vspace{-0.2cm}

\subsubsection{Model Transfer API}
Most open-source libraries that utilize LLM's are compatible with Hugging Face's model format. Since JORA uses JAX for its training procedure, the caveat is incompatibility with the popular libraries. To overcome this limitation, we provide a simple script to convert models trained using our library to the Hugging Face format. Listing~\ref{lst:conversion} provides a description of the conversion script usage.

JORA builds on a Llama-2 implementation in JAX which uses jit and vmap. Llama-2 is a GPT based model. It uses the decoder component of the transformer architecture to produce text autoregressively. Since transformer models consist of multi-headed self-attention, the memory used at the inference stage scales quadratically with the input sequence length. This is a significant drawback for RAFT since augmenting a prompt with retrieved-context adds to the sequence length. As such, one of the aims of our library is to assuage the memory utilization requirements by efficiently distributing memory usage across GPU resources. 

\begin{lstlisting}[label=lst:conversion, caption=Hugging Face conversion script can be invoked from the command-line. The converted model can be used with other Hugging Face compatible libraries such as LangChain.]
SYNOPSIS
    huggingface_merger.py HUGGINGFACE_PATH JAX_PATH SAVE_PATH

POSITIONAL ARGUMENTS
    HUGGINGFACE_PATH
        Type: str
        path to the HuggingFace llama model
    JAX_PATH
        Type: str
        path to LoRA parameters fine-tuned by JORA
    SAVE_PATH
        Type: str
        path to save the updated HuggingFace llama model
\end{lstlisting}
\vspace{-0.2cm}

For our implementation of LoRA, we follow the suggestions presented by \citet{hu2021lora}, i.e., the query and value attention weights are enhanced. Specifically, the approach suggests that the computation, $W_0x + b_0$, can be tuned through $W_0x + b_0 + BAx$ where $W_0$ are subset of the models weights, $B$, $A$ are the trainable countports of $W_0$ added by LoRA, $W_0, BA \in \mathbb{R}^{m \times n}$, $A \in \mathbb{R}^{r \times n}$, $B \in \mathbb{R}^{m \times r}$, and $r << m,n$. 

Here, $B$ and $A$ are the trainable weights. $W_0$ and $b_0$ represent the weights and biases of a specific neural network component. Composing the trainable parameters to lower rank values significantly reduces the total parameters involved in backpropagation. Generative models are trained to predict the next token, given past tokens auto-regressively. Thus, the objective, $\mathcal{L}$, of the LLM is to reduce the discrepancy between the next predicted token $\hat{y}_{t+1}$ and the next ground truth token $y_{t+1}$, given the past tokens in the ground truth sequence, $y_0^t$. Consequently, the trained language model predicts the next token, given the past predicted tokens, $\hat{y}_0^t$.

For our implementation of LoRA, we add the LoRA parameters to the original weights as highlighted in Equation~\ref{eq1}. The values of $B$ and $A$ are initialized from zeros and normal sampling, respectively.

\begin{equation} \label{eq1}
\begin{split}
Output & = W_0x + b_0 + BAx \\
 & = (W_0 + BA) x + b_0
\end{split}
\end{equation}
    
JORA parallelizes all parameters of the Llama model using JAX's positional sharding module. Transformers inherently support distributed computations through the use of parallel decoder blocks. Llama-2 consists of several layers of parallel decoder blocks. We utilize the inherent design and shard on the decoder axis. Projection and Embedding layers are sharded on the non-sequential dimension to avoid variation due to the input. 

\subsubsection{Library Usage}
One of the core aims of JORA is to make fine-tuning easily accessible to the end-user. Compared to Hugging Face, JORA significantly reduces the lines of code to get started. In addition, JORA provides a GUI for fine-tuning LLMs. The following code can be used to fine-tune a model with minimal changes to default training parameters:
\begin{lstlisting}[language=python]
from jora import train_lora, ParallamaConfig, generate_alpaca_dataset

config = ParallamaConfig(
        MODEL_SIZE=model_size, 
        JAX_PARAMS_PATH=jax_path,
        LLAMA2_META_PATH=hf_path)
dataset = generate_alpaca_dataset(dataset_path, 'train', config)
train_lora(config, dataset, checkpoint_path)
\end{lstlisting}

Alternatively, the GUI can set the fine-tuning parameters and training. Fig.~\ref{fig:gui} shows the interface for the GUI. It can be invoked with the following command:
\begin{lstlisting}
python -m jora.gui
\end{lstlisting}

\begin{figure}
    \centering
    \includegraphics[width=\columnwidth]{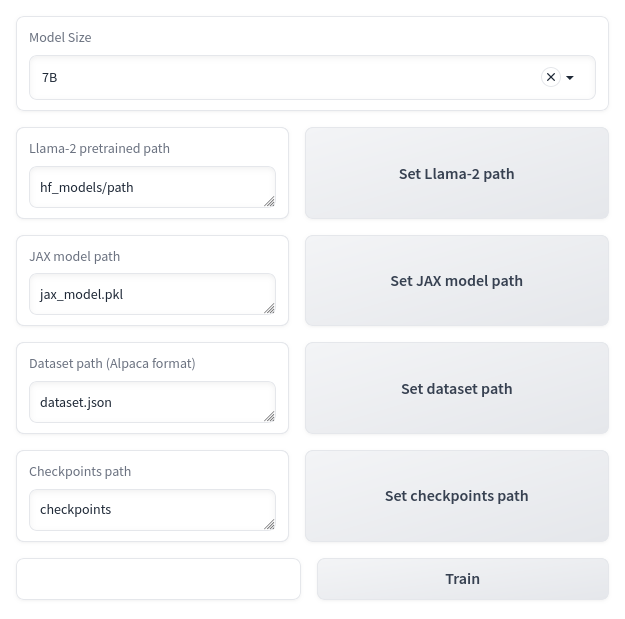}
    \caption{JORA provides a simple GUI for fine-tuning.}
    \label{fig:gui}
\end{figure}

\section{Experiments}

\begin{table*}[]
\centering
\resizebox{.8\textwidth}{!}{%
\begin{tabular}{@{}clrrr@{}}
\cmidrule(l){2-5}
\multicolumn{1}{l}{} &
  \multicolumn{1}{c}{\textbf{GPUs}} &
  \multicolumn{1}{c}{1} &
  \multicolumn{1}{c}{2} &
  \multicolumn{1}{c}{4} \\ \midrule
\multirow{2}{*}{\begin{tabular}[c]{@{}c@{}}Hugging Face PEFT w/\\ Microsoft DeepSpeed ZeRO-3\end{tabular}} &
  \textbf{Mem (MB)} &
  \textbf{\begin{tabular}[c]{@{}r@{}}20645.2\\ (39.81)\end{tabular}} &
  \begin{tabular}[c]{@{}r@{}}23056 / 23024\\ (14.63 / 29.29)\end{tabular} &
  \begin{tabular}[c]{@{}r@{}}23978 / 23921 / 23463 / 23397\\ (47.87 / 50.39 / 31.96 / 17.46)\end{tabular} \\
 &
  \textbf{Performance (secs)} &
  \begin{tabular}[c]{@{}r@{}}4.56\\ (0.04)\end{tabular} &
  \begin{tabular}[c]{@{}r@{}}2.81\\ (0.02)\end{tabular} &
  \begin{tabular}[c]{@{}r@{}}5.45\\ (0.09)\end{tabular} \\ \cmidrule(r){1-1}
\multirow{2}{*}{JORA (Ours)} &
  \textbf{Mem (MB)} &
  \begin{tabular}[c]{@{}r@{}}23102\\ (0.00)\end{tabular} &
  \textbf{\begin{tabular}[c]{@{}r@{}}16068 / 16008\\ (0.00 / 0.00)\end{tabular}} &
  \textbf{\begin{tabular}[c]{@{}r@{}}11460 / 11448 / 11448 / 11400\\ (0.0 / 0.00 / 0.00 / 0.00)\end{tabular}} \\
 &
  \textbf{Performance (secs)} &
  \textbf{\begin{tabular}[c]{@{}r@{}}0.19\\ (0.00)\end{tabular}} &
  \textbf{\begin{tabular}[c]{@{}r@{}}0.79\\ (0.00)\end{tabular}} &
  \textbf{\begin{tabular}[c]{@{}r@{}}0.44\\ (0.00)\end{tabular}} \\ \bottomrule
\end{tabular}%
}
\caption{JORA shows significant improvement w.r.t. Hugging Face implementation of PEFT paired with DeepSpeed for parallelization. JORA uses tensor-parallelism to distribute memory allocation for parameters across GPU resources. The number in the brackets denotes the standard deviation across five runs.} 
\label{tbl:performance}
\vspace{-0.5cm}
\end{table*}

We measure the improvement introduced by JORA in terms of memory utilization and computation speed, conducting experiments using Hugging Face/DeepSpeed for comparison. Our setup consists of a system with 4 x A100 with 40GB of VRAM each, an AMD EPYC 75F3 32-core Processor, and 512GB of RAM. The GPUs are cross-connected using NVLink. All experiments use brain floating point for parameter precision for a fair comparison.

\subsection{Memory Utilization Analysis}
We compare the memory utilization of our implementation with that of the Hugging Face trainer using Accelerate and PEFT. Our implementation is adapted from the examples in the official Hugging Face PEFT library, which uses Accelerate and DeepSpeed for parallel computation. Through parallelization, several parameters are replicated across multiple GPUs. As such, the total memory utilized by parallel training is greater than that used in a single GPU setting. However, the advantage of multi-GPU training is that the memory used by each GPU individually is less than that used in single-GPU training. JAX pre-allocates memory to avoid fragmentation, which makes measuring active allocation a challenge. 
For memory utilization analysis, we override this behavior by setting the XLA\_PYTHON\_CLIENT\_ALLOCATOR environment variable to `platform.' This environment variable informs JAX to allocate and deallocate memory as needed but impacts performance. Thus, for the performance evaluation, we use the default configuration.

For parallel training, DeepSpeed distributes parameters using data parallelism. Thus, though a single sample cannot be distributed, multiple samples can be aggregated, improving performance. Thus, JORA is beneficial since it allows a single lengthy sequence to backpropagate across multiple GPUs. Table~\ref{tbl:performance} shows that JORA uses less memory per resource as the number of resources increases. The only case where Hugging Face/DeepSpeed consumes lower memory is where only one GPU is available.

\subsection{Computation Time Comparison}
We also measure computation time using the same RAFT dataset for the Hugging Face and JORA implementations over iterations of 1, 2, and 4 GPUs. Table~\ref{tbl:performance} presents these results. JORA shows consistently better performance than Hugging Face implementation, with JORA implementation being over 12 times faster than the baseline with 4 GPUs. Since DeepSpeed used data parallelism, we observe a performance impact in multi-GPU settings, with the bottleneck being the slowest GPU/sample for backpropagation. 
In addition to improved performance, since JORA uses JAX's jit functionality to run compiled computations, the performance of the implementation shows more consistency. We observe a computation performance drop between single and multiple GPUs. This drop could be attributed to cross-GPU communication overhead. 


\section{An Example Usage Scenario}
JORA is designed to aid in RAFT. In this section, we demonstrate a RAFT use case by fine-tuning it on a social media dataset to help LLMs enable social-context understanding. The purpose of RAG is to add additional context to a prompt by searching for knowledge and adding additional information. For RAFT, data can be created based on retrieved knowledge. The LLM learns to generate the retrieved answer based on the context since the key rationale is held back. A simple example is a database query, which corresponds to a process that may be taken to produce an output by evaluating the database. If the query is not provided but rather a natural language equivalent is provided, the LLM must learn the heuristics represented by the hidden query. 

Since prompt tuning is insufficient for models to develop social-context understanding~\cite{gandhi2023understanding}, we use a fine-tuning process consisting of two phases to add knowledge to an LLM. Both phases of fine-tuning use PEFT. For our problem setting, rather than just predicting the following words, we aim to gain an understanding of the relation across different comments in a social media session. For instance, a comment in a social media session may target the previous comment, the original post that spawned the session, or some comment in the middle of the discourse. To glean insight into the target of the comment in terms of its context, reasoning between the structure of the conversation is critical. Unfortunately, the LLM pre-training does not consider these relationships specifically, and there is no public data related to reasoning at the comment level in social media discourse. Thus, we rely on other general-purpose structured data as a surrogate to learn structure and reasoning. We use the WikiTableQuestions~\cite{pasupat2015compositional} dataset to infuse structural intelligence into the model. This dataset consists of various independent tables, questions based on one of the tables, and a corresponding answer. To answer these questions, using the data in the input table is vital. Some answers require aggregate reasoning. 

\begin{table}[]
\centering
\resizebox{\columnwidth}{!}{%
\begin{tabular}{@{}lccc@{}}
\cmidrule(l){2-4}
                  & \multicolumn{1}{l}{\textbf{Target Post}} & \multicolumn{1}{l}{\textbf{Reply Post}} & \multicolumn{1}{l}{\textbf{p(Reply | Target)}} \\ \cmidrule(l){2-4} 
\textbf{7B}       & 0.082                                    & 0.153                                   & 0.643                                          \\
\textbf{13B}      & 0.159                                    & 0.200                                   & {\ul 0.815}                                    \\
\textbf{7B-RAFT}  & {\ul 0.865}                              & {\ul 0.541}                             & 0.558                                          \\
\textbf{13B-RAFT} & \textbf{0.971}                           & \textbf{0.847}                          & \textbf{0.855}                                 \\ \bottomrule
\end{tabular}%
}
\caption{The veracity of the directionality identification improves with the RAFT fine-tuning phases w.r.t. the baselines. Given the conversation as context, the values represent the accuracy of detecting the respective posts.}
\label{tab:directionality_analysis}
\vspace{-0.5cm}
\end{table}
\raggedbottom

For the directionality analysis task~(which post is targeted by another comment in the same session), we leveraged a corpus of 4chan threads~\cite{papasavva2020raiders}. This dataset consists of $\sim$$3$ million threads and $\sim$$100$ million posts. Because 4chan allows its users to tag whom they reply to, we use this data as the ground truth for directionality information. We examine whether our RAFT phases improve (i) the model's ability to detect the post we are targeting for behavior comprehension and (ii) the model's ability to distinguish who is being targeted by the poster. 4chan allows posters to mention more than one comment as the target of the reply. Here, we consider the model successful if one of the multiple comments is identified. Table~\ref{tab:directionality_analysis} shows the result of our experiment. The RAFT model significantly improves performance over the pre-trained counterparts. This illustrates the application of RAFT to improve LLM performance in social media analysis. \textit{Social media conversation threads can provide important context but they can span large sequences. JORA helps in the training process here by splitting a discourse sequence's computation tensors across multiple GPUs. This is not possible using HuggingFace/Deepspeed because Data-Parallelism in these frameworks distributes the workload between different data instances rather than dividing the computation for a single data instance among multiple accelerators.}


\section{Conclusion}
This paper presents JORA, a JAX-based library for Retrieval Augment fine-tuning of Llama-2 models. JORA provides convenient functions for data manipulation and training. In addition, it implements best practices for memory efficient and performant training. By using a combination of LoRA, tensor-parallelism, and jit, JORA can significantly improve memory efficiency and computation time over a distributed environment compared to Hugging Face/DeepSpeed. Finally, JORA can export trained models to the popular Hugging Face model format for downstream usage with other Hugging Face-compatible libraries. 

\bibliography{bibliography}

\end{document}